\documentclass[conference]{IEEEtran}
\IEEEoverridecommandlockouts

\usepackage{times}

\usepackage[numbers,sort&compress]{natbib}
\usepackage{multicol}
\usepackage{graphicx}
\usepackage{epsfig} 
\usepackage{times} 
\usepackage{amsmath} 
\usepackage{amssymb}  
\usepackage{soul}

\usepackage{graphicx}
\usepackage{algorithm} 
\usepackage{algpseudocode} 
\usepackage{multirow} 
\usepackage{caption}
\usepackage{makecell}
\usepackage{booktabs}
\usepackage{xcolor}
\usepackage{cuted} 

\definecolor{mylinkcolor}{RGB}{40, 115, 201}
\definecolor{mycitecolor}{RGB}{71, 191, 38}
\definecolor{DGreen}{RGB}{107, 190, 35}
\definecolor{mygreen}{RGB}{106, 168, 79}
\definecolor{myblue}{RGB}{60, 120, 216}
\definecolor{myred}{RGB}{204, 0, 0}

\usepackage[bookmarks=true, colorlinks=true, citecolor=mycitecolor,linkcolor=mylinkcolor,urlcolor=mycitecolor]{hyperref}

\begin{document}

\title{TACO: Temporal Consensus Optimization for Continual Neural Mapping}

\author{\authorblockN{Xunlan Zhou$^{\ast}$\thanks{$^{\ast}$Equal contribution.}}

\authorblockA{School of Intelligence Science\\ and Technology\\
Nanjing University\\
Email: wyattzhouxl@smail.nju.edu.cn}
\and
\authorblockN{Hongrui Zhao$^{\ast}$}

\authorblockA{Department of Aerospace Engineering\\
University of Illinois Urbana-Champaign\\
Email: hongrui5@illinois.edu}

\and
\authorblockN{Negar Mehr}

\authorblockA{Department of Mechanical Engineering\\
University of California Berkeley\\
Email: negar@berkeley.edu}
}



%

\maketitle

\begin{strip}
  \begin{minipage}{\textwidth}\centering
    \vspace{-40pt}
        \includegraphics[width=\textwidth]{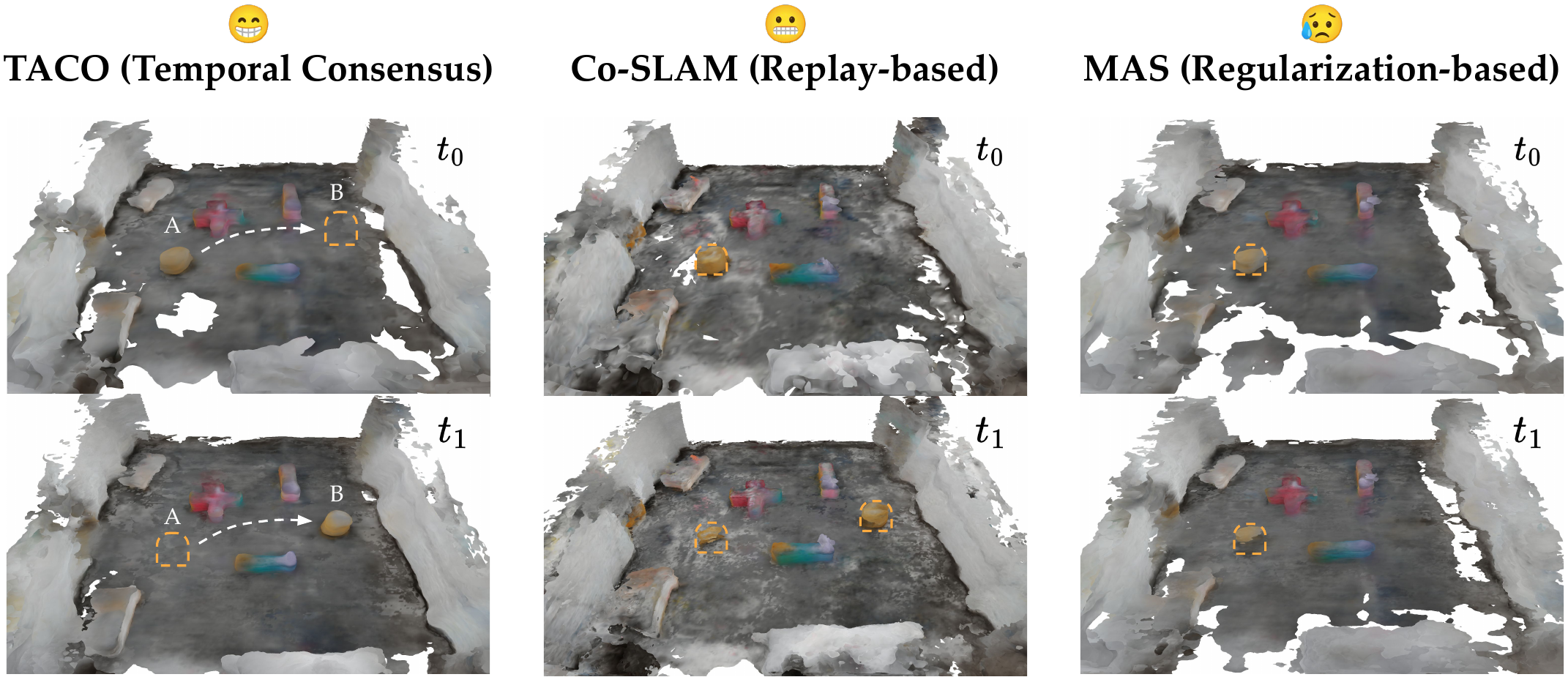}  
    \captionof{figure}{\textbf{Comparison of continual neural mapping under environment changes.} The yellow stool was moved from $A$ to $B$.
\textbf{Left:} TACO (Temporal Consensus Optimization) revises outdated regions while preserving consistent geometry and yields an accurate and up-to-date map.
\textbf{Middle:} Replay-based methods (e.g., Co-SLAM) preserve high-quality reconstructions overall, but their reliance on replaying past observations can lead to persistent ``ghost'' artifacts when scene geometry changes.
\textbf{Right:} Regularization-based methods (e.g., MAS) over-constrain the optimization, unable to adapt to the changing environment.}

    \label{fig:teaser}
  \end{minipage}
\end{strip}


\begin{abstract}
Neural implicit mapping has emerged as a powerful paradigm for robotic navigation and scene understanding. 
However, real-world robotic deployment requires continual adaptation to changing environments under strict memory and computation constraints, which existing mapping systems fail to support. 
Most prior methods rely on replaying historical observations to preserve consistency and assume static scenes. 
As a result, they cannot enable continual learning in dynamic robotic tasks.
To address these challenges, we propose \textbf{TACO} (\textbf{T}empor\textbf{A}l \textbf{C}onsensus \textbf{O}ptimization), a replay-free framework for continual neural mapping. 
We reformulate mapping as a temporal consensus optimization problem, where we treat past model snapshots as temporal neighbors. 
Intuitively, our approach resembles a model consulting its own past knowledge.
We update the current map by enforcing weighted consensus with historical representations. 
Our method allows reliable past geometry to constrain optimization while permitting unreliable or outdated regions to be revised in response to new observations. 
TACO achieves a balance between memory efficiency and adaptability without storing or replaying previous data. 
Through extensive simulated and real-world experiments, we show that TACO robustly adapts to scene changes, and consistently outperforms other continual learning baselines. The code can be
found at \href{https://iconlab.negarmehr.com/TACO}{\textcolor{mycitecolor}{https://iconlab.negarmehr.com/TACO}}
\end{abstract}

\IEEEpeerreviewmaketitle

\section{Introduction}

Neural implicit representation has become an important paradigm for robotic dense scene mapping \citep{mildenhall2021nerf,wang2021neus,yariv2021volume,yu2022monosdf}. 
Unlike traditional discrete map representations \citep{6162880,niessner2013real,whelan2015elasticfusion}, they encode scene geometry and appearance as continuous functions, which yields compact and expressive scene models with high reconstruction fidelity and low memory cost.
Recent work demonstrates that neural implicit maps support direct optimization from streaming RGB-D observations and enable online dense mapping at real-time rates \citep{zhu2022nice,wang2023co,sucar2021imap}. 
Despite their success, most existing online neural mapping systems rely on storing and replaying past observations or keyframes to stabilize training and prevent the forgetting of the previously visited regions. 
However, in dynamic real-world environments, past and new observations often conflict. 
Training on such inconsistent data degrades map quality and negatively impacts downstream tasks which require up-to-date environmental information.
In addition to keyframe replay, methods such as Continual Neural Mapping (CNM) \cite{yan2021CNM} and UNIKD \cite{guo2024unikd} draw inspiration from continual learning literature \citep{de2021continual,parisi2019continual,shin2017continual,guo2024unikd,robins1995catastrophic} to prevent forgetting. 
However, these approaches often struggle to adapt to dynamic scenes, as they overly constrain updates to prevent deviation from the prior model.

\begin{figure}[t!]
    \centering
    \includegraphics[width=0.98\linewidth]{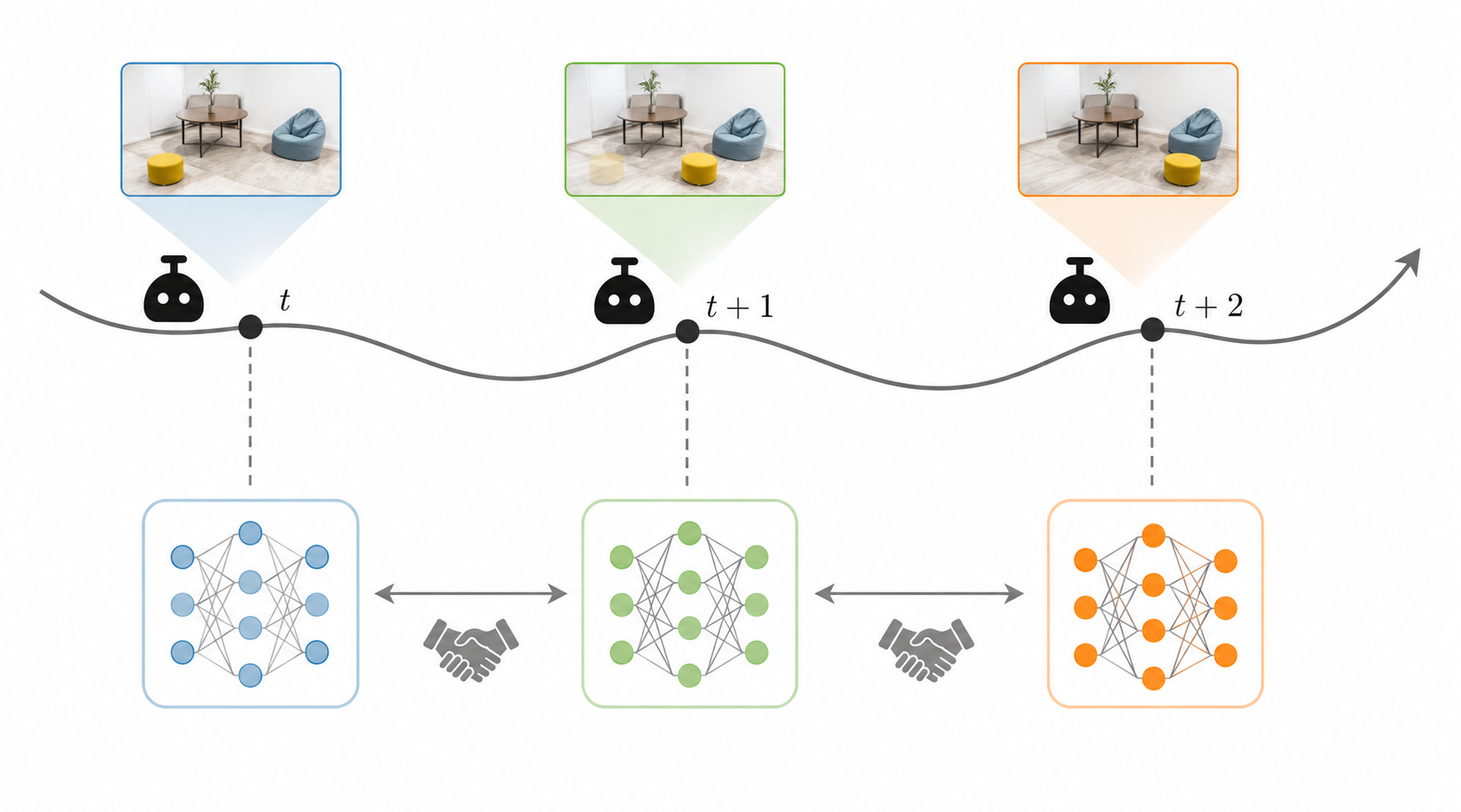}
    \caption{\textbf{Intuition of temporal consensus.} A robot receives sequential visual observations over time, while neural map snapshots are encouraged to reach consensus across adjacent time steps. This allows the map to preserve reliable past knowledge while adapting to newly observed scene changes.}
    \label{fig:tacopipeline}
\end{figure}

To address these challenges, we introduce TACO, a Temporal Consensus Optimization framework for continual neural mapping. 
We preserve previous model parameters as ``historical snapshots''. 
Upon receiving new observations, a \emph{temporal consensus} process enforces consistency between the current model parameters and these snapshots.
We also incorporate an importance-aware weighting scheme to enable a more adaptive consensus enforcement.
Intuitively, our weighting allows the current model to align with reliable aspects of its past states while remaining free to revise regions that conflict with new observations. 
As illustrated in Fig.~\ref{fig:tacopipeline}, a robot receives sequential observations over time, while the corresponding neural map snapshots are encouraged to reach consensus across adjacent time steps.

Our approach avoids catastrophic forgetting and allows adaptation to the changing environment, as illustrated in Fig.~\ref{fig:teaser}. 
Our formulation enables continual adaptation under streaming observations \emph{without relying on data replay} and remains well-suited for long-term online mapping under realistic resource constraints. 
Through extensive experiments, we demonstrate that we consistently produce more accurate and up-to-date reconstructions under environment changes compared to existing baselines.

In summary, our key contributions in this paper are:
\begin{enumerate}
    \item We propose TACO, a Temporal Consensus Optimization framework that reformulates continual neural mapping as a consensus problem over time.
    \item We introduce an importance-aware temporal consensus mechanism that dynamically balances stability and adaptation: we preserve reliable past knowledge, while updating regions that conflict with new observations.
    \item We validate TACO through extensive simulation and real-world robotic experiments. We show that TACO adapts to dynamic scene changes while state-of-the-art methods fail. Our approach mitigates forgetting under realistic online and memory-constrained conditions.
\end{enumerate}

\section{Related Works}

\textbf{Continual Learning.}
Continual learning studies how a model acquires knowledge over time from sequential data streams while preserving useful past information and adapting to new data \citep{parisi2019continual,wang2024comprehensive,de2021continual,ryu2025curricullm}. A widely studied subset is incremental learning which considers settings where training data arrive sequentially and only a subset of observations is available at each step \citep{ade2013methods,de2021continual}. Most incremental learning methods address catastrophic forgetting under the assumption of a stationary task or data distribution and typically fall into three categories \citep{de2021continual}: data replay \citep{chaudhry2018efficient,klein2007kr,lopez2017gradient,rebuffi2017icarl,rolnick2019experience,shin2017continual}, parameter regularization \citep{aljundi2018MAS,lin2021barf,rannen2017encoder,kirkpatrick2017ewc}, and parameter isolation \citep{aljundi2017expert,fernando2017pathnet,mallya2018packnet,xu2018reinforced}. Representative examples include replay-based strategies such as PTAM with keyframe replaying \cite{klein2007kr}, regularization-based approaches like MAS \cite{aljundi2018MAS} and EWC \cite{kirkpatrick2017ewc}, parameter isolation methods such as PackNet \cite{mallya2018packnet}, and knowledge distillation techniques such as POD \cite{douillard2020podnet}, AFC \cite{kang2022class}, and UNIKD \cite{guo2024unikd}. Some of these strategies have also been applied to neural implicit representations. Examples include Continual Neural Mapping (CNM) \cite{yan2021CNM} for neural surface fields, CLNeRF \cite{cai2023clnerf} for NeRF-based novel view synthesis, and UNIKD \cite{guo2024unikd}, which leverages uncertainty-filtered distillation to mitigate forgetting in incremental neural implicit reconstruction.

While effective at preserving previously reconstructed content, these approaches primarily focus on consistency with past model states. 
In neural mapping systems, this emphasis often manifests as aligning with earlier reconstructions, even when new sensor observations indicate geometric changes. 
As a result, the map can retain outdated structure and fail to reflect scene evolution over time. 
Our work instead considers a broader continual mapping setting \citep{parisi2019continual,de2021continual}, where the objective extends beyond preventing forgetting to updating historical geometry in response to new observations without requiring access to past data.

\textbf{Consensus Optimization for Neural Mapping.}
Consensus optimization provides an approach to reconcile conflicting information across multiple learners by combining local updates with consensus constraints \citep{nedic2009distributed}. 
In neural mapping, this paradigm has been explored primarily in multi-robot settings, where independent agents observe different parts of the scene and must agree on a shared representation. Prior work applies consensus-based optimization to neural implicit mapping in order to maintain geometric consistency across agents under limited or unreliable communication \citep{yu2022dinno,asadi2024dinerf,deng2024macim,zhao2024distributed,dou2025openmulti,zhao2025ramen,zhao2025udon}. 
Although originally developed for spatially distributed systems, these methods reveal a broader insight: \emph{consensus mechanisms, especially when weighted by importance, provide an effective means of resolving conflicting information without requiring access to raw data}. 
In this work, we build on this insight by shifting the notion of consensus from the spatial domain to the temporal domain. 
Rather than coordinating multiple agents, we propose to coordinate multiple model states over time, which enables continual neural mapping that updates outdated geometry while preserving reliable structure.

\textbf{Finding the Important Parameters in Neural Implicit Maps.}
Quantifying the importance of model parameters in neural implicit maps is relevant to tasks such as view selection, refinement, and artifact suppression, as well as to continual and incremental optimization. 
Much of the existing literature approaches this problem from an uncertainty quantification perspective. 
Classical techniques such as deep ensembles \cite{sunderhauf2022density} and Bayesian neural networks \cite{shen2021stochastic} provide principled uncertainty quantification approaches. Alternative methods estimate spatial uncertainty by perturbing inputs or parameters \citep{goli2024bayes,Yan_2023_ICCV}; these approaches primarily capture output variability and do not explicitly characterize the relative importance of individual learnable parameters. 
More recent works utilize computationally efficient uncertainty proxies to accelerate real-time mapping.
RAMEN \cite{zhao2025ramen} estimates parameter relevance using frequency-based statistics over multi-resolution hash grids to modulate consensus in distributed neural mapping. 
UNIKD \cite{guo2024unikd} incorporates uncertainty estimates into a distillation framework to filter unreliable knowledge when transferring information across incremental training stages. 
In contrast, our work adopts an output-sensitivity-based importance formulation inspired by Memory Aware Synapses (MAS) \cite{aljundi2018MAS}. 
We quantify parameter importance through the sensitivity of network outputs to parameter perturbations.
This measure reflects the temporal reliability of learned representations, and we use it to regulate continual optimization without introducing additional network components or auxiliary inference procedures.

\section{Preliminaries}
In this section, we introduce the core building blocks and notations used throughout the paper.
We first review Co-SLAM \cite{wang2023co}, a real-time neural implicit mapping framework that serves as the optimization backbone of our approach.
We then provide an overview of the method of multipliers \cite{ADMM,CADMM}, which provides the algorithmic foundation of our work for later incorporating temporal consensus into neural mapping.

\subsection{Neural Implicit Mapping with Co-SLAM}
Co-SLAM represents a scene using a neural implicit map composed of three components: 
a multi-resolution hash-based feature grid \cite{instaNGP}, denoted as 
$V_\alpha = \{ V_\alpha^l \}_{l=1}^L$, 
a geometry decoder $F_\tau$, and a color decoder $F_\phi$. 
The feature grid contains $L$ resolution levels with geometrically increasing resolutions, where higher levels encode finer geometric details. 
We denote the learnable parameters of the neural implicit map as 
$\Theta = \{ \alpha, \tau, \phi \}$, 
where $\alpha$ parameterizes the hash grid, and $\tau$ and $\phi$ parameterize the geometry and color decoders, respectively.

Given a 3D point with world coordinate ${x}$, the neural implicit map predicts its truncated signed distance function (SDF) value $s$ and RGB color ${c}$.
The SDF value encodes the signed distance to the closest surface, with the sign indicating whether the point lies inside or outside the surface.
To improve geometric smoothness, one-blob positional encoding \cite{oneblob} $\gamma(\cdot)$ is applied to ${x}$.
Using the encoded coordinate $\gamma({x})$ and interpolated feature vectors $V_\alpha({x})$, the geometry decoder outputs a latent feature ${h}$ and the SDF value:
\begin{equation}
    F_\tau \left( \gamma({x}), V_\alpha({x}) \right) \to ({h}, s).
    \label{eq:geometry}
\end{equation}
The color decoder then predicts the RGB value:
\begin{equation}
    F_\phi \left( \gamma({x}), {h} \right) \to {c}.
    \label{eq:color}
\end{equation}

Co-SLAM renders RGB and depth images by volume rendering along camera rays.
Given camera intrinsics $K$ and pose $P$, each image pixel with coordinates $(u,v)$ corresponds to a ray with origin ${o}$ and direction ${r} = P K^{-1}(u,v,1)^\top$.
Along each ray, $M$ sample points ${x}_i = {o} + d_i {r}$ are evaluated, where $d_i$ denotes the distance from the camera origin.
For each ${x}_i$, the neural implicit map predicts $(s_i, {c}_i)$ using \eqref{eq:geometry} and \eqref{eq:color}.
The rendered pixel color $\hat{{c}}$ and depth $\hat{d}$ are computed as weighted averages:
\begin{subequations}
\begin{align}
    \hat{{c}} &= \frac{1}{\sum_{i=1}^M w_i} \sum_{i=1}^M w_i {c}_i,\label{eq:rgb} \\
    \hat{d} &= \frac{1}{\sum_{i=1}^M w_i} \sum_{i=1}^M w_i d_i,\label{eq:depth}
\end{align}
\end{subequations}
where $w_i$ is the weight quantifying the contribution of point $i$ to the final rendering \cite{neuralRGBD}.
We truncate the SDF value $s$ to $tr$ to exclude the points far from the surface, as these regions provide little information for reconstruction.
Then, we compute the weights as
\begin{equation}
    w_i = \sigma\!\left(\frac{s_i}{tr}\right)\sigma\!\left(-\frac{s_i}{tr}\right),
\end{equation}
with $\sigma(\cdot)$ denoting the Sigmoid function. 
This formulation assigns high weights to samples whose signed distance $s_i$ is close to zero, which corresponds to points near the estimated surface intersection along the ray, while suppressing contributions from samples in free space.
The sigmoid functions provide a smooth truncation controlled by $tr$, avoiding hard thresholds and ensuring stable gradients during optimization by softly limiting the influence of samples far from the surface.

The neural implicit map is optimized by minimizing the discrepancy between rendered RGB-D images and observations collected by the robot.
Let $R$ denote the incoming RGB-D frames.
The training loss is defined as
\begin{equation}
    L^{obj}({\Theta}, R) = L_{rgb} + L_d + L_{sdf} + L_{fs} + L_{smooth},
    \label{eq:obj}
\end{equation}
where $L_{rgb}$ and $L_d$ measure the photometric and depth reconstruction errors, $L_{sdf}$ supervises SDF predictions, $L_{fs}$ constrains points far from surfaces, and $L_{smooth}$ encourages geometric smoothness.
We refer readers to \cite{wang2023co} for detailed definitions of each loss term.

\subsection{Constrained Optimization with the Method of Multipliers}
In this section, we review the method of multipliers \cite{CADMM} for solving constrained optimization problems, which serves as a key algorithmic building block in our approach.
Consider the following optimization problem
\begin{equation}
    \label{eq:decentralized}
    \min_{\Theta} \; L^{obj}(\Theta, R), \; \text{s.t.} \; A \Theta = z,
\end{equation}
where the linear equality constraint $A \Theta = z$ defines the feasible parameter space. 
Here, $A$ is the constraint matrix and $z$ is the target vector. 
The neural implicit map parameters are flattened into a vector, denoted as $\Theta$,     which serves as the optimization variable.

To solve \eqref{eq:decentralized}, we can construct the augmented Lagrangian of the problem:
\begin{equation}
\label{eq:augLa}
\mathcal{L}^{a} = L^{obj}({\Theta}, R) + (A \Theta - z)^\top p + \frac{\rho}{2}  \Vert A \Theta - z\Vert_2^2,
\end{equation}
where $p$ is the vector of dual variables, and $\rho>0$ is a penalty scalar parameter controlling the strength of enforcing constraint. 
We apply dual ascent to minimize the augmented Lagrangian \eqref{eq:augLa}, which iteratively updates the dual and the primal variables.
At iteration $k$, the updates take the form:
\begin{subequations}
    \begin{align}
        \Theta^{k} &= \arg\min_{\Theta} L^{obj}({\Theta}, R) + (A \Theta - z)^\top p^{k-1}  \nonumber \\ 
        & + \frac{\rho}{2}  \Vert A {\Theta} - z\Vert_2^2, \label{eq:primal_update} \\
        p^{k} &= p^{k-1} + \rho \left( A \Theta^{k} - z \right). \label{eq:dual_update}
    \end{align}
\label{eq:ADMM}
\end{subequations}
The update in \eqref{eq:primal_update} is referred to as the \emph{primal update}, while \eqref{eq:dual_update} corresponds to the \emph{dual update}.
Intuitively, the dual variable acts as accumulated disagreement signals that record how much our primal variable $\Theta$ violates the constraint over past iterations. 
The dual variable drives subsequent updates to correct these discrepancies.
Since $\Theta$ is the neural network parameters, we only approximate the primal update using a small number of stochastic gradient descent steps.
In the next section, we will discuss how we develop our temporal consensus formulation based on this framework.

\section{Temporal Consensus Optimization for Continual Neural Mapping}

Now, we explain the main idea behind TACO.
We propose to view consensus as enforcing agreement between the current model and its historical snapshots (model parameters from the past). 
From this perspective, continual neural mapping can be viewed as optimizing a current neural implicit map from streaming observations while retaining a set of frozen historical snapshots that encode past knowledge.

\subsection{Problem Formulation}
Let ${\Theta}_t$ denote the neural implicit map parameters obtained by optimizing on the incoming observations $R_t$ at time step $t$.
We consider an online continual mapping setting in which observations arrive sequentially, and the model is updated incrementally using only the current data stream, without storing or revisiting past RGB-D frames.

This temporal setting gives rise to two competing requirements that a continual neural mapping system must balance:
\begin{enumerate}
    \item \emph{Adaptation to scene changes:}
    Historical models may encode outdated or incorrect geometry when the environment evolves, and indiscriminately enforcing consistency with such information can impede the incorporation of new observations.

    \item \emph{Retention of valid historical knowledge:}
    At the same time, historical models often contain accurate information about regions of the environment that remain unchanged.
    An effective continual learning strategy should preserve such reliable knowledge to avoid unnecessary relearning.
\end{enumerate}

To address these competing requirements, we formulate continual neural mapping as a temporal consensus optimization problem.
We force the current model toward agreement with a set of frozen historical snapshots $\{{\Theta}_{t-k}\}$, where ${\Theta}_{t-k}$ denotes parameters obtained at earlier time steps.
We also introduce importance-aware weighting to modulate the strength of consensus to preserve reliable historical knowledge while enabling adaptation to newly observed regions.

\subsection{Importance-Weighted Temporal Consensus}
For simplicity, we consider only the current model $\Theta_t$ and its immediate predecessor $\Theta_{t-1}$. 
We aim to update the neural map to incorporate the new observation $R_{t}$ while ensuring that it does not deviate significantly from the previous state $\Theta_{t-1}$.
To enforce this, we define a consensus target $z_{t,t-1}(\Theta_t)$ 
\begin{align}
&z_{t,t-1}(\Theta_t)
=
\left( {W}_t(\Theta_t)  + {W}_{t-1}(\Theta_t) \right)^{-1} \nonumber \\
&\left( W_t(\Theta_t) \Theta_t + {W}_{t-1}(\Theta_t) \Theta_{t-1} \right),
\label{eq:temporal_z}
\end{align}
where $W_t(.)$ and ${W}_{t-1}(.)$ are square weight matrices encoding parameter-wise importance of $\Theta_t$ and $\Theta_{t-1}$.
We note that $W_t, W_{t-1}$ and $z_{t,t-1}$ are functions of $\Theta_t$.
We depend $W_{t-1}$ on $\Theta_t$ due to the normalization process.
We compute $W_t, W_{t-1}$ using the gradient information of $\Theta_t$.
We will discuss the computations of $W_t, W_{t-1}$ in detail in the next section.
Constraint \eqref{eq:temporal_z} computes a parameter-wise weighted average of the current and historical models, where parameters with higher weights exert a stronger influence on the consensus target.
As a result, important historical parameters are preserved, while parameters associated with lower weights are allowed to adapt to new observations.

In this setting, our optimization objective is:
\begin{equation}
\label{eq:temporal_consensus}
\min_{\Theta_t} \; L^{obj}({\Theta_t}, R_t), \; \text{s.t.} \; \Theta_t = z_{t,t-1}(\Theta_t).
\end{equation}
Here, $L^{obj}(\Theta_t, R_t)$ denotes the reconstruction loss with the new observations.
Following the process described in the preliminaries section, we next obtain the augmented Lagrangian of \eqref{eq:temporal_consensus} as
\begin{align}
\mathcal{L}^{a} = & \; 
L^{obj}(\Theta_t, R_t) + \left( \Theta_t - z_{t,t-1}(\Theta_t)  \right)^\top p \notag\\ 
&+ \frac{\rho}{2}\| \Theta_t - z_{t,t-1}(\Theta_t)  \|_{2}^2,
\label{eq:temporal_aug}
\end{align}
where $\rho$, a user-picked hyperparameter, controls the strength of temporal consensus, and $p$ is the dual variable.

We now solve this optimization problem \eqref{eq:temporal_aug} iteratively by applying the method of multipliers. 
We update the primal and dual variables for $K$ iterations.
As mentioned in \eqref{eq:temporal_z}, $W_t, W_{t-1}$ and $z_{t,t-1}$ are functions of $\Theta_t$.
Thus, we also need to update them when the value of $\Theta_t$ is changing at each iteration.
After the dual update of the previous iteration $k-1$, we utilize the information of $ \Theta_t^{k-1}$ to compute $W_t^{k-1}$ and $W_{t-1}^{k-1}$. 
Then, with $W_t^{k-1}, W_{t-1}^{k-1}$ and $\Theta_t^{k-1}$, we can get $z_{t,t-1}^{k-1}$:
\begin{align}
&z_{t,t-1}^{k-1}(\Theta_t^{k-1})
=
\left( W_t^{k-1}(\Theta_t^{k-1})  + W_{t-1}^{k-1}(\Theta_t^{k-1}) \right)^{-1} \nonumber \\
&\left( W_t^{k-1}(\Theta_t^{k-1}) \Theta_t^{k-1} + W_{t-1}^{k-1}(\Theta_t^{k-1}) \Theta_{t-1} \right),
\label{eq:z_update}
\end{align}
At the current iteration $k$, the state of the neural map, $\Theta_t^{k}$, is obtained by the primal update:
\begin{align}
\Theta_t^{k}
&=
\arg\min_{\Theta_t}
\; L^{obj}(\Theta_t, R_t)
+(\Theta_t - z_{t,t-1}^{k-1}(\Theta_t^{k-1}))^\top {p}^{k-1} \notag \\
& + \frac{\rho}{2}
\|
\Theta_t - z_{t,t-1}^{k-1}(\Theta_t^{k-1})
\|_2^2.
\label{eq:temporal_primal}
\end{align}
We then update the dual variable:
\begin{equation}
p^{k} = p^{k-1} + \rho ( \Theta_t^{k} - z_{t,t-1}^{k-1}(\Theta_t^{k-1}) ).
\label{eq:temporal_dual}
\end{equation}
This dual update penalizes deviations from the importance-weighted temporal consensus target while allowing reliable historical parameters to exert stronger influence on the optimization.

After the method of multipliers iterative updates, we set the next state of the neural map to be $\Theta_t = \Theta_t^{K}$, where $K$ is the last iteration.
In the next section, we will discuss the importance weight matrix in detail.

\subsection{Importance Estimation for Temporal Consensus}

To construct importance-aware weights for temporal consensus optimization, we estimate a parameter-wise importance that controls how strongly each parameter is constrained over time.
In neural implicit mapping, parameters contribute unequally to representing past scene geometry, and uniform temporal constraints can either impede adaptation or preserve outdated structure.
Our importance measure captures each parameter’s contribution to the learned geometry, protecting reliable regions while allowing flexibility where the scene has changed.



To measure how strongly model predictions depend on individual parameters, we first define a proxy objective on the rendered outputs:
\begin{equation}
\mathcal{L}^{\mathrm{proxy}}_{s}
=
\frac{1}{|\mathcal{S}_s|}
\sum_{r \in \mathcal{S}_s}
\Big(
\|\hat{{c}}_{s,r}\|_2^2
+
\|\hat{d}_{s,r}\|_2^2
\Big),
\label{eq:mas_uncertainty_proxy}
\end{equation}
where $\mathcal{S}_s$ denotes the set of rays or pixels sampled at method of multiplier iteration $s$, and $\hat{{c}}_{s,r}$ and $\hat{d}_{s,r}$ are the predicted color and depth defined in Eqs.~\eqref{eq:rgb} and \eqref{eq:depth}.
This proxy objective captures the overall magnitude of the model’s predictions and serves as a task-agnostic signal for probing output sensitivity without requiring ground-truth supervision.

We then estimate parameter importance by accumulating the sensitivity of this proxy objective with respect to the model parameters over iterations:
\begin{align}
{u}_t^k(\Theta_t^k)
&=
\sum_{s=1}^{k}
\left|
\frac{\partial \, \mathcal{L}^{\mathrm{proxy}}_s}{\partial {\Theta}_t^s}
\right|
= 
\left|
\frac{\partial \, \mathcal{L}^{\mathrm{proxy}}_k}{\partial {\Theta}_t^k}
\right| + u_t^{k-1}.
\label{eq:mas_uncertainty_u}
\end{align}
Here, $s$ indexes iterations (up to current iteration $k$), and ${\Theta}_t^s$ denotes the optimization variable at the iteration $s$.
We compute $\nabla_{{\Theta}_s}\mathcal{L}^{\mathrm{proxy}}_s$ via backpropagation and accumulate its element-wise absolute value.

Intuitively, ${u}_t^k$ measures how sensitive the model’s predictions are to perturbations of each parameter across past iterations.
Parameters with larger ${u}_t^k$ have consistently exerted stronger influence on the rendered outputs and are therefore treated as more important, while parameters with smaller values are allowed to adapt more freely.


Since accumulated importance can grow unboundedly over time, it cannot be directly used as a consensus weight in our optimization.
Moreover, independently normalizing the importance vectors of the current and historical models (e.g., to $[0,1]$) destroys the relative magnitude information across parameters.

To preserve relative importance while maintaining numerical stability, we adopt a relative linear scaling strategy to construct temporal consensus weights.
Using \eqref{eq:mas_uncertainty_u}, we obtain the importance vector ${u}_t^k$. 
We also have $u_{t-1}$ saved alongside the frozen historical snapshot ${\Theta}_{t-1}$.
We then compute their joint importance
${u}_{\mathrm{sum}} ^k:= {u}_t^k + {u}_{t-1}$.
Rather than imposing a fixed range, we normalize the overall scale of the weights by aligning their average magnitude with the penalty parameter $\rho$. 
Specifically, we define a global scaling factor
\begin{equation}
    \epsilon^k(\Theta_t^k) = \frac{\rho}{\mathrm{avg}({u}_{\mathrm{sum}}^k)}.
\end{equation}
By normalizing with the average over all parameters of ${u}_{\mathrm{sum}}^k$, we preserve the relative importance structure across parameters while preventing the absolute weight magnitudes from becoming arbitrarily large or small.

Then, we can construct our weight matrices as the following diagonal matrices
\begin{equation}
W_t^k(\Theta_t^k)= \mathrm{diag}(\epsilon^k \cdot  {u}_t^k),
\qquad
{W}_{t-1}^k(\Theta_t^k) = \mathrm{diag}(\epsilon^k \cdot {u}_{t-1}).
\label{eq:taco_weight}
\end{equation}
where $\mathrm{diag}(\cdot)$ denotes the diagonal matrix formed by placing the entries of a vector on the main diagonal.


\begin{figure*}[t!]
    \centering
    \includegraphics[width=0.9\textwidth]{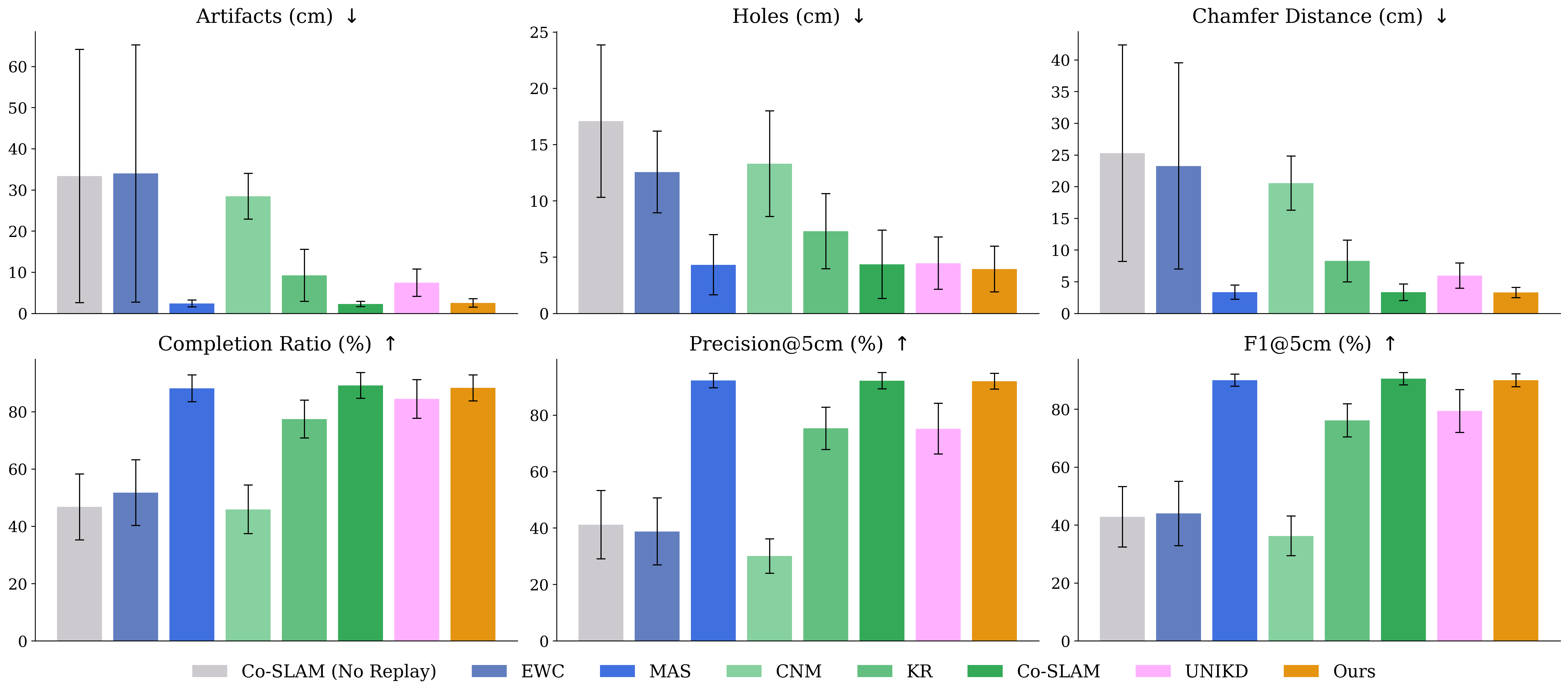}
    \caption{Quantitative results on eight Replica scenes under static settings.
    We report reconstruction accuracy and completeness metrics, including Artifacts, Holes, Chamfer Distance, Completion Ratio, Precision@5cm, and F1@5cm.
    TACO achieves performance comparable to Co-SLAM with replay and matches or outperforms all continual learning baselines across all metrics.}
    \label{fig:replica}
\end{figure*}

Note that during early mapping stages, some parameters may exhibit non-zero importance values due to noise, initialization effects, or indirect parameter coupling, even though they do not yet encode stable scene structure. Applying temporal consensus uniformly to such parameters can introduce unnecessary constraints in the early stages of the training. To make the importance-weighted consensus more robust in these cases, we incorporate a simple masking mechanism on the temporal consensus weights.
Specifically, given a threshold $\beta > 0$, we suppress temporal constraints on parameters whose associated weights fall below this threshold:
\begin{equation}
W_{t-1}^{(i,i)} =
\begin{cases}
W_{t-1}^{(i,i)}, & W_{t-1}^{(i,i)} \ge \beta, \\
0, & W_{t-1}^{(i,i)} < \beta.
\end{cases}
\label{eq:weight_mask}
\end{equation}

This masking operation effectively disables temporal consensus for parameters with extremely low importance, allowing unconstrained adaptation in poorly observed regions, while preserving strong temporal constraints where sufficient evidence has been accumulated. 


\section{Experiments and Results}



We evaluate \textbf{TACO} in both simulation and real-world settings considering static and dynamic scenes.
Evaluating both regimes serves two complementary goals: static scenes allow us to measure how well we preserve previously learned geometry over time without forgetting; dynamic scenes introduce controlled scene changes to evaluate whether we can adapt to new observations while retaining reliable historical structure.

We conduct static-scene experiments on two standard RGB-D benchmarks: 
\textbf{Replica} \cite{replica19arxiv} which provides high-fidelity synthetic indoor scenes with accurate ground-truth geometry, and 
\textbf{ScanNet} \cite{dai2017scannet} which captures large-scale real-world indoor environments with sensor noise and partial observations.

To study continual adaptation under dynamic conditions, we construct dynamic benchmarks in both simulation and the real world.
In simulation, we base our dynamic experiments on the \textbf{Habitat Synthetic Scenes Dataset} \cite{khanna2024habitat} and use \textbf{Habitat-Sim} \citep{habitat19iccv,szot2021habitat,puig2023habitat3} to introduce controlled object-level scene changes. 
We further collect a \textbf{real-world dynamic dataset} using a turtlebot.
We deliberately modify the scene to validate that TACO remains effective when the environment changes.

We compare TACO against a representative set of continual learning baselines, all implemented within the Co-SLAM framework for a controlled comparison:
\begin{itemize}
    \item \textbf{Parameter regularization methods:} \textbf{EWC} \cite{kirkpatrick2017ewc} and \textbf{MAS} \cite{aljundi2018MAS}, which mitigate forgetting by penalizing changes to parameters deemed important for past observations.
    \item \textbf{Replay-based methods:} \textbf{CNM} \cite{yan2021CNM} and \textbf{KR} \cite{klein2007kr}, which preserve historical information via explicit data replay. For KR, we use a replay buffer of 10 keyframes \cite{sucar2021imap}.
    \item \textbf{Distillation-based method:} \textbf{UNIKD} \cite{guo2024unikd}, which leverages uncertainty-filtered knowledge distillation to reduce forgetting during incremental updates.
\end{itemize}

\begin{figure*}[t!]
    \centering
    \includegraphics[width=0.9\textwidth]{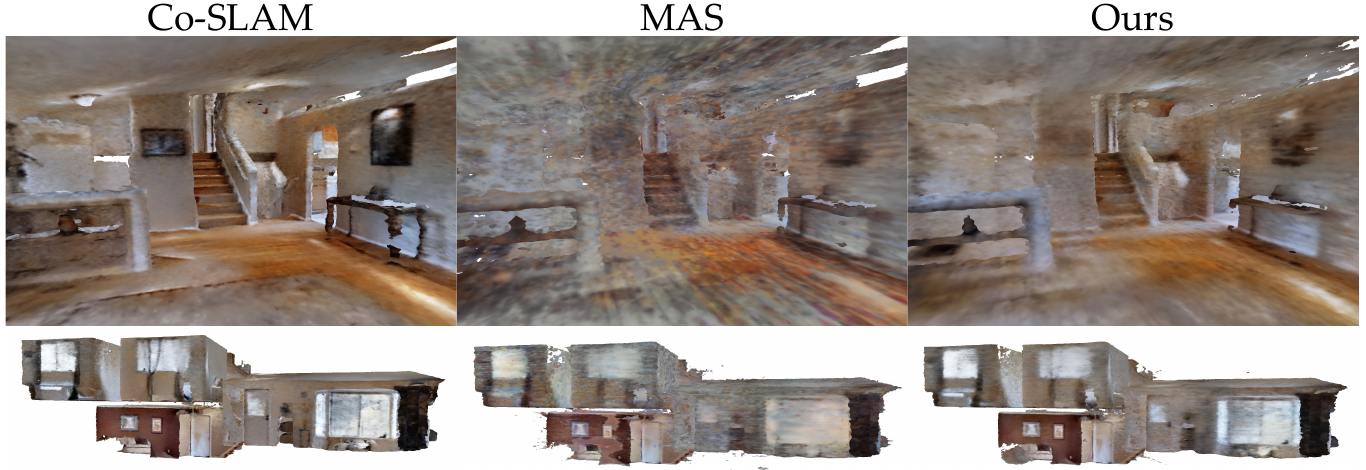}
    \caption{Qualitative reconstruction results on a representative Gibson scene. TACO achieves geometry quality comparable to the replay-based baseline while producing less textural artifacts than MAS.}
    \label{fig:gibson}
\end{figure*}

We also report results for \textbf{Co-SLAM with replay} and \textbf{Co-SLAM without replay} as reference points.
Note that Co-SLAM with replay represents an upper bound as it optimizes over both past and current observations via a replay buffer, while Co-SLAM without replay serves as a lower bound, updating only on recent observations and exposing the severity of catastrophic forgetting.

All our continual learning baselines operate in a strict streaming setting: observations are processed sequentially, past data are not revisited unless explicitly allowed by the method (e.g., replay-based baselines).

To quantitatively evaluate reconstruction accuracy, we adopt a set of standard 3D geometric metrics widely used in prior neural implicit mapping works \citep{sucar2021imap,zhu2022nice,zhao2025ramen,guo2024unikd}.
These metrics jointly assess surface accuracy, completeness, and artifact suppression.
We have provided their detailed definitions and evaluation protocols in Appendix \ref{Appendix A}.

\subsection{Results on Static Simulated Scenes}

We evaluate all our methods on static scenes to assess their ability to mitigate catastrophic forgetting when the environment remains unchanged.
As shown in Fig. \ref{fig:replica}, TACO achieves top-tier performance on the Replica benchmarks, closely matching the upper bound given by Co-SLAM with replay and performing on par with or better than the strongest continual learning baseline, MAS.
More details, including results on ScanNet, can be found in the Appendix \ref{Appendix B}.
These results indicate that importance-weighted temporal consensus effectively preserves global geometry over time without revisiting past observations.

To further test scalability under static conditions, we evaluate the top performing methods, {TACO}, {MAS}, and {Co-SLAM}, on a larger and more structurally complex environment from the Gibson dataset \cite{xiazamirhe2018gibsonenv}.
Compared to Replica and ScanNet, Gibson scenes cover substantially larger spatial extents and exhibit increased geometric complexity.
As shown in Fig.~\ref{fig:gibson}, TACO produces more complete and smoother geometric reconstructions than MAS, with fewer discontinuities and less mesh fragmentation.
Although both TACO and MAS exhibit color inconsistency relative to replay-based Co-SLAM due to the absence of explicit appearance modeling, TACO maintains coherent geometry without relying on replay.
These results demonstrate that temporal consensus scales effectively to larger static environments.

For the subsequent evaluations on dynamic scenes, we only compare TACO to {Co-SLAM with replay} and {MAS}, which are the strongest baseline in the static settings.

\subsection{Results on Dynamic Simulated Scenes}

We further evaluate continual neural mapping under dynamic environment changes using simulated scenes with staged object motion.
Unlike static reconstruction settings, dynamic scenes require the model to adapt to newly observed geometry while preserving previously reliable map content, making them a direct test of continual adaptation.

\begin{figure*}[t!]
    \centering
    \includegraphics[width=\textwidth]{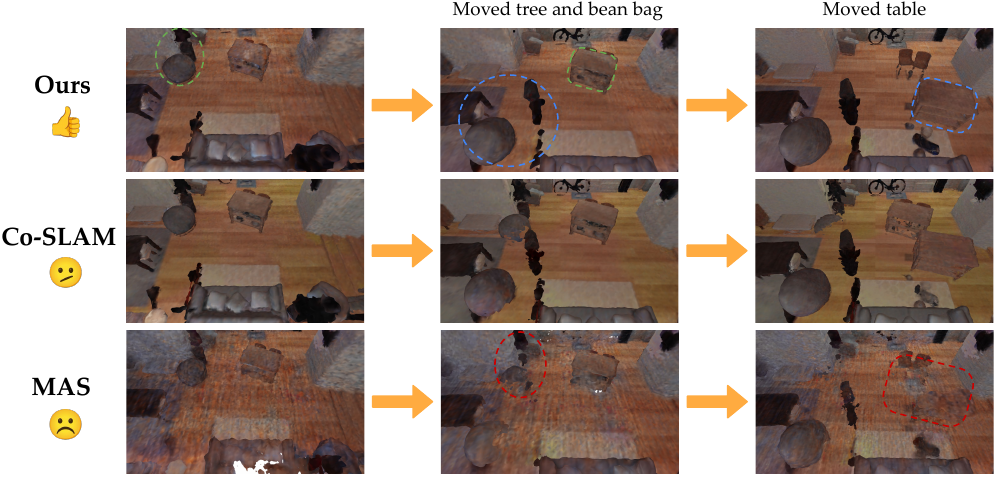}
    \caption{\textbf{Dynamic scene reconstruction under staged object motion.}
TACO adapts to scene changes without introducing artifacts in stable regions, whereas replay- and regularization-based methods produce ghosting and mesh tearing, respectively.
Dashed boxes highlight objects affected by staged motion and the resulting reconstruction behavior: \textcolor{mygreen}{green boxes} mark objects that will move in the next stage, \textcolor{myblue}{blue boxes} mark objects whose positions have changed in the current stage, and \textcolor{myred}{red boxes} highlight failure regions with severe artifacts.}
    \label{fig:stimulation}
\end{figure*}

To construct this evaluation, we build dynamic datasets based on the Habitat Synthetic Scenes Dataset.
Using Habitat-Sim, we generate simulated environments in which scene geometry is modified in a staged manner, allowing us to isolate and analyze the behavior of different continual mapping strategies under conflicting observations.
Specifically, the dynamic sequence consists of two stages, as illustrated in Fig.~\ref{fig:stimulation}.
In the first stage, a beanbag sofa and a potted tree are relocated to new positions, while the remaining scene structure remains unchanged.
In the second stage, the table is moved to a different location, and the tableware originally stored on the built-in lower shelf is disturbed and scattered onto the carpet, which introduces additional fine-grained geometric changes.
Throughout both stages, large portions of the scene remain static. 

Fig.~\ref{fig:stimulation} illustrates qualitative reconstruction results across stages.
TACO accurately captures the staged scene modifications while maintaining consistent reconstruction of unchanged regions.
After each change, outdated geometry is correctly revised, and the updated map reflects the current object configuration without introducing artifacts in previously stable areas.
This demonstrates that temporal consensus enables selective adaptation rather than uniform preservation of past states. In contrast, MAS, as a regularization-based method, enforces strong parameter constraints to preserve prior knowledge.
When confronted with conflicting observations caused by object motion, this over-constraining behavior prevents proper geometric revision.
As a result, mesh tearing and fragmented surfaces emerge near both the original and updated object locations.
Replay-based methods such as Co-SLAM exhibit a different failure mode.
By explicitly replaying past observations, Co-SLAM enforces consistency with both historical and current geometry.
When objects are moved, this leads to simultaneous reconstruction at both the old and new positions, producing persistent ghost artifacts as a result.
While replay mitigates catastrophic forgetting in static settings, it becomes detrimental under environment changes where past observations are no longer valid.

Table~\ref{tab:dynamic_results} reports quantitative results on the dynamic simulated scenes.
TACO consistently outperforms MAS across all metrics. It produces fewer reconstruction artifacts and achieves higher completion ratio and F1@5cm.
These results demonstrate that the proposed temporal consensus mechanism is effective not only for preserving past geometry but also for adapting to dynamic scene changes.

\begin{table}[h]
\centering
\small
\caption{Quantitative results on dynamic simulated scenes. TACO outperforms the regularization-based MAS baseline, producing fewer reconstruction artifacts and achieving higher completion ratio and F1 score.}
\label{tab:dynamic_results}
\begin{tabular}{lccc}
\toprule
Method & Artifacts $\downarrow$ & Completion Ratio $\uparrow$ & F1@5cm $\uparrow$ \\
\midrule
MAS   & 8.53 & 71.41 & 73.43 \\
TACO  & \color{DGreen}\textbf{6.38} & \color{DGreen}\textbf{74.37} & \color{DGreen}\textbf{78.05} \\
\bottomrule
\end{tabular}
\end{table}

Overall, these qualitative and quantitative results highlight the limitations of replay- and regularization-based continual mapping under dynamic scenes.
By treating past representations as temporally weighted constraints rather than immutable supervision, TACO achieves robust adaptation to scene changes while preserving consistent geometry.

\subsection{Continual Mapping in Dynamic Real-World Environments}

We further evaluate continual neural mapping performance in a real-world setting with controlled scene changes.

We first consider a simple scenario where we move a single object—the yellow stool highlighted in Fig.~\ref{fig:teaser}—from position $A$ at time $t_0$ to position $B$ at time $t_1$ while keeping the rest of the scene unchanged. As shown in Fig.~\ref{fig:teaser}, different methods exhibit distinct behaviors after the relocation.
Replay-based methods such as Co-SLAM reconstruct the stool at both its original and new locations. It produces persistent artifacts because replay enforces consistency with historical observations even when they no longer reflect the current scene.
Regularization-based methods such as MAS strongly penalize parameter updates associated with past observations, which over-constrains optimization and leads to distorted or fragmented geometry in regions affected by the conflict.
In contrast, our method selectively relaxes constraints on parameters linked to outdated geometry while preserving consistency in unchanged regions.
As a result, we reconstruct the stool only at its current location, without ghost artifacts or geometric breakage.


\begin{figure*}[t!]
    \centering
    \includegraphics[width=0.9\textwidth]{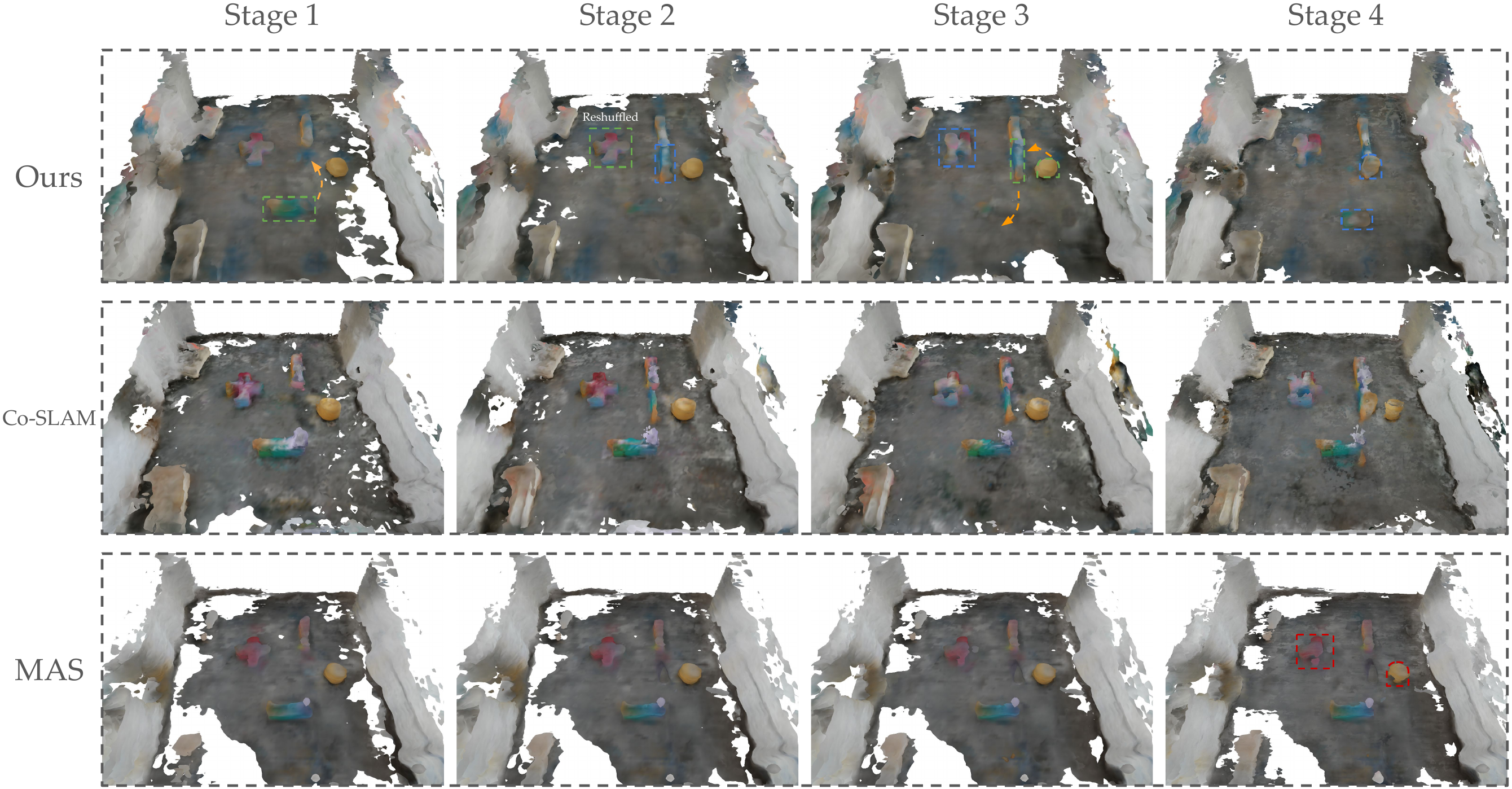}
    \caption{Additional hardware experiment with multiple sequential scene changes. TACO consistently adapts to changing environments, outperforming the baseline methods. Dashed boxes follow the same annotation scheme as in Fig.~\ref{fig:stimulation}: \textcolor{mygreen}{green boxes} mark objects that will move in the next stage, \textcolor{myblue}{blue boxes} mark objects whose positions have changed in the current stage, and \textcolor{myred}{red boxes} highlight failure regions with severe artifacts.}
    \label{fig:hardware2}
\end{figure*}

We further test continual mapping under more challenging real-world dynamics by introducing multiple sequential scene changes.
As illustrated in Fig.~\ref{fig:hardware2}, this experiment proceeds through four stages.
In the first stage, we move a row of cubes to new locations. In the second stage, we reshuffle a larger set of cubes into a different configuration. In the third stage, we move two cubes back to their original positions and relocate the yellow stool to the positions previously occupied by those cubes.
Across these stages, the environment undergoes repeated and heterogeneous changes that induce accumulating conflicts between historical and current observations.

Across all stages, we adapt consistently to the evolving scene by revising outdated geometry while preserving stable regions and producing coherent reconstructions throughout the sequence. In contrast, Co-SLAM accumulates severe ghost artifacts as changes progress, and MAS fails to adapt meaningfully under repeated changes as it produces torn and fragmented geometry in regions affected by object motion.

Overall, these real-world experiments demonstrate that we robustly balance stability and adaptivity across real-world dynamics, and validate temporal consensus as an effective mechanism for replay-free continual neural mapping in dynamic environments.

\subsection{Analysis of Memory Usage}

We analyze the CPU memory footprint of representative methods as CPU memory is a critical constraint for real-world robotic deployment.
We report \emph{resident set size} (RSS), which measures the amount of physical CPU memory actively occupied by a process during execution, excluding disk-backed virtual memory.
Unlike GPU memory, CPU RSS directly reflects the on-board memory pressure faced by robotic systems and thus provides a practical indicator of deployability.

\begin{table}[h!]
\centering
\caption{CPU memory footprint (RSS) comparison in MB.}
\begin{tabular}{lc}
\toprule
Method & CPU RSS (MB) $\downarrow$ \\
\midrule
Co-SLAM & 4165.1 \\
KR (Replay-based) & 1912.4 \\
UNIKD (Knowledge Distillation) & 1945.1 \\
Online MAS (Regularization) & 1800.6 \\
Ours & 1864.5 \\
\bottomrule
\end{tabular}
\label{tab:cpu_memory}
\end{table}

As shown in Table~\ref{tab:cpu_memory}, full replay (Co-SLAM) incurs substantial memory overhead due to storing historical observations.
In contrast, TACO achieves a CPU memory footprint comparable to replay-based, distillation-based, and regularization-based baselines, while reducing memory usage by more than 50\% relative to full replay.

\section{Conclusion} 
\label{sec:conclusion}

We introduced TACO, a replay-free framework for continual neural implicit mapping that formulates online mapping as an importance-aware temporal consensus optimization problem. By treating historical models as temporally weighted constraints rather than immutable supervision, TACO enables selective revision of outdated geometry while preserving consistent structure in unchanged regions. Our experiments in both simulated and real-world environments demonstrate that TACO robustly adapts to scene changes while avoiding ghost artifacts and over-constrained updates common to replay- and regularization-based methods. This work highlights temporal consensus as a principled mechanism for scalable long-term neural mapping in dynamic robotic environments.

\section*{Acknowledgments}

This work was supported in part by NSF grants ECCS-2438314 (CAREER Award) and CNS-2529645. 
This work was also supported by the UC Berkeley Center for Digital Assets (CDA).
This research was made possible by GPU resources provided via the NVIDIA Academic Grant.
\bibliographystyle{plainnat}
\bibliography{references}

\clearpage

\appendix

\subsection{Quantitative Metrics for Geometry Evaluation}  \label{Appendix A}

To assess the accuracy and completeness of reconstructed geometry, we employ a set of 3D quantitative metrics commonly used in prior neural implicit mapping works
\citep{sucar2021imap,zhu2022nice,zhao2025ramen,guo2024unikd}, with minor renaming for clarity.
Together, these metrics capture complementary aspects of geometric fidelity, surface completeness, and artifact suppression.

\begin{itemize}
    \item \textbf{Artifacts (cm)} measure the average distance from reconstructed surface points to the ground-truth geometry, reflecting the presence of spurious or floating artifacts. Lower values indicate fewer artifacts.
    
    \item \textbf{Holes (cm)} compute the average distance from ground-truth surface points to the reconstructed mesh, quantifying missing or unreconstructed regions. Lower values indicate better coverage.
    
    \item \textbf{Completion Ratio (\%)} reports the percentage of ground-truth points whose nearest reconstructed point lies within a 5\,cm threshold. This metric serves as our primary measure of surface completeness.
    
    \item \textbf{Chamfer Distance (CD)} is defined as the symmetric average distance between reconstructed and ground-truth point sets, jointly reflecting geometric accuracy and surface completeness. Lower values are better.
    
    \item \textbf{Precision@5\,cm} measures the fraction of reconstructed surface points whose nearest ground-truth point lies within a 5\,cm threshold, penalizing spurious geometry and floating artifacts. Higher values are better.
    
    \item \textbf{F1@5\,cm} is the harmonic mean of Precision@5\,cm and Recall@5\,cm at a 5\,cm threshold, providing a balanced metric that captures both geometric fidelity and scene coverage.
\end{itemize}

\subsection{Results on Static Simulated Scenes}  \label{Appendix B}

We evaluate all methods on static scenes to establish a controlled setting for assessing reconstruction quality and catastrophic forgetting in the absence of scene changes.
Fig.~\ref{fig:replica} reports results on eight Replica scenes across all metrics.
Among parameter-regularization baselines, {EWC} collapses severely, exhibiting large artifacts and Chamfer errors and substantially reduced completion, precision, and F1 scores, indicating that naive stability constraints are insufficient for continual neural mapping.
In contrast, {MAS} performs strongly and remains competitive with the best methods across most metrics.
Replay-based methods ({CNM}, {KR}) partially recover performance—{KR} consistently outperforms {CNM}—but both fall short of the top tier in terms of accuracy and coverage.
The distillation-based baseline {UNIKD} improves over replay and regularization failures and achieves solid reconstruction quality, yet remains below the best-performing methods overall.

Across all Replica scenes, we achieve top-tier performance on every metric, closely matching {Co-SLAM} with replay and exhibiting performance comparable to or better than {MAS}.
In particular, TACO maintains low artifact and Chamfer distance while preserving high completion, precision@5cm, and F1@5cm, demonstrating that importance-weighted temporal consensus preserves global geometry without revisiting past observations.
Compared to explicit replay ({CNM}, {KR}) and knowledge distillation ({UNIKD}), TACO achieves a more favorable accuracy–completeness trade-off, reducing spurious geometry without sacrificing surface coverage.

\begin{figure}[t!]
    \centering
    \includegraphics[width=\linewidth]{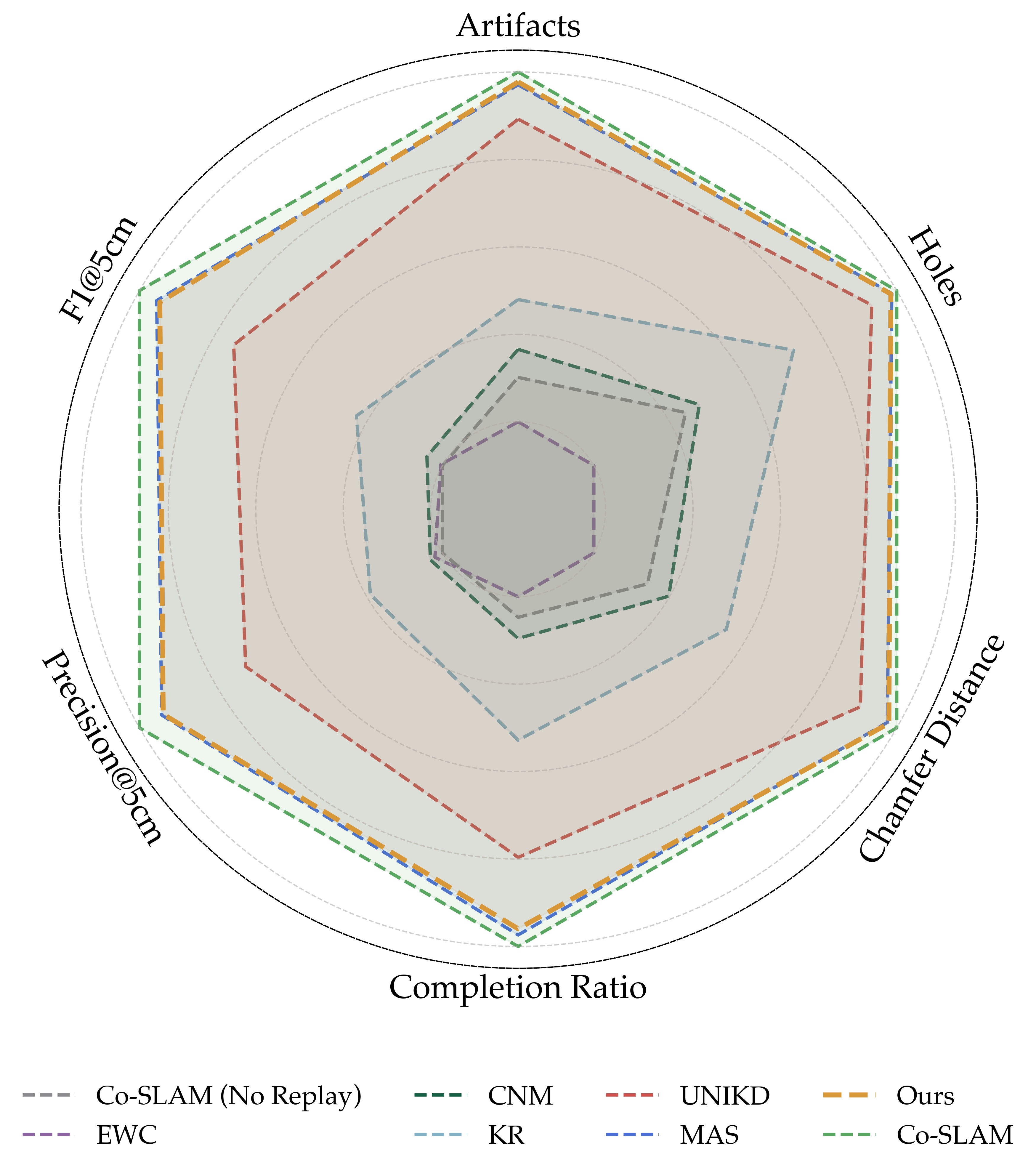}
    \caption{Quantitative Results on ScanNet}
    \label{fig:scannet_radar}
\end{figure}

We further provide quantitative evaluation on two real-world ScanNet scenes (\texttt{scene0000\_00} and \texttt{scene0673\_05}), summarized by the radar plots in Fig.~\ref{fig:scannet_radar}.
The overall performance trends closely mirror those observed on Replica.
Parameter-regularization baselines degrade substantially across all metrics, reflecting limited robustness to real-world sensor noise and partial observations.
Replay-based and distillation-based methods recover partial performance but remain uneven across accuracy and completeness dimensions.
In contrast, TACO consistently lies near the upper envelope of the radar plots, closely aligning with {Co-SLAM} and matching the strongest continual learning baseline across all metrics.

\begin{figure*}[t!]
    \centering
    \includegraphics[width=\textwidth]{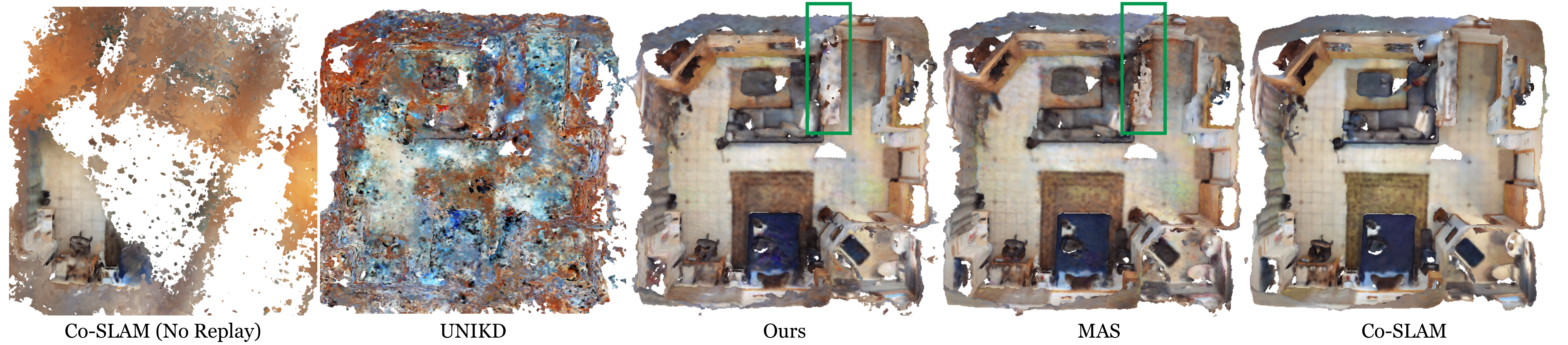}
    \caption{Visualization on ScanNet \texttt{scene0000\_00}.}
    \label{fig:scannet}
\end{figure*}

We next visualize reconstruction quality on ScanNet.
As shown in Fig.~\ref{fig:scannet}, results on \texttt{scene0000\_00} reveal consistent qualitative behaviors under real-world sensing noise and incomplete coverage.
{Co-SLAM} without replay fails to preserve coherent geometry and produces fragmented reconstructions, while {UNIKD} largely maintains scene structure but exhibits severe color inconsistency.
Regularization-based methods such as {MAS} preserve the overall layout but show noticeable geometric breakage in several regions.
In contrast, TACO produces stable and coherent reconstructions and more completely recovers the table region highlighted in the figure.

\subsection{Ablation on Mask Mechanism}  \label{Appendix C}

We conduct an ablation study to analyze the effect of the mask mechanism used in temporal consensus weighting.
The mask is designed to suppress unreliable parameter updates during early stages of mapping, where non-informative gradients may arise due to noise, initialization effects, or indirect coupling within the network.
Setting the mask to zero disables this mechanism, corresponding to importance-based temporal consensus without masking.

\begin{table}[h]
\centering
\caption{Ablation study on the mask mechanism for temporal consensus weighting on \texttt{room1}.}
\label{tab:mask_ablation}
\begin{tabular}{lcc}
\toprule
Method & Chamfer Distance (CD) $\downarrow$ & F1@5\,cm $\uparrow$ \\
\midrule
Ours ($\tau=10^{-2}$) & 2.87 & 89.96 \\
Ours ($\tau =10^{-3}$) & 2.21 & 92.69 \\
Ours ($\tau=10^{-4}$) & \textbf{2.05} & \textbf{93.65} \\
Ours ($\tau=10^{-5}$) & 2.12 & 93.20 \\
Ours ($\tau=0$)              & 2.17 & 93.06 \\
\bottomrule
\end{tabular}
\end{table}

Table~\ref{tab:mask_ablation} reports results on \texttt{room1} using Chamfer Distance (CD) and F1@5\,cm.
Introducing the mask consistently improves reconstruction quality compared to the no-mask baseline.
In particular, moderate masking strengths (e.g., $1\mathrm{e}{-4}$) achieve the best trade-off between geometric accuracy and surface completeness, yielding the lowest CD and highest F1 score.
When the mask is too strong, performance degrades due to overly aggressive suppression of parameter updates, while disabling the mask entirely leads to slightly inferior results.

These results indicate that the proposed mask mechanism plays a critical role in stabilizing importance-aware temporal consensus, enabling robust continual adaptation while avoiding the influence of spurious early-stage gradients.

We further observe that the choice of the masking threshold $\tau$ is coupled with the penalty parameter $\rho$, which controls the overall strength of temporal consensus.
Larger values of $\rho$ amplify the influence of consensus terms, requiring a correspondingly higher threshold $\tau$ to avoid over-constraining early-stage or unreliable parameters, while smaller $\rho$ values allow for weaker masking without destabilizing optimization.

\subsection{Ablation on the Number of Temporal Snapshots}  \label{Appendix D}

We conduct an ablation study to examine the effect of using multiple historical snapshots for temporal consensus optimization.
Specifically, we vary the number of snapshots $K$ jointly participating in temporal consensus while keeping all other components fixed.

\begin{table}[h]
\centering
\caption{Ablation on the number of temporal snapshots $K$ used for temporal consensus on \texttt{room1}.}
\label{tab:k_snapshot_ablation}
\begin{tabular}{lcc}
\toprule
$K$ (number of snapshots) & Chamfer Distance (CD) $\downarrow$ & F1@5\,cm $\uparrow$ \\
\midrule
$K=1$ & \textbf{2.50} & \textbf{93.65} \\
$K=2$ & 2.62 & 91.53 \\
$K=3$ & 2.85 & 90.29 \\
$K=4$ & 3.51 & 87.60 \\
\bottomrule
\end{tabular}
\end{table}

Table~\ref{tab:k_snapshot_ablation} reports results on Replica \texttt{room1}.
Using a single historical snapshot ($K=1$) yields the best overall performance, achieving the lowest Chamfer Distance and highest F1 score.
As $K$ increases, performance degrades consistently across both metrics.

This behavior reflects the trade-off between stability and adaptability in temporal consensus.
In dynamic mapping scenarios, older snapshots are more likely to encode outdated geometry.
Including multiple historical models introduces conflicting constraints that bias optimization toward obsolete scene states, making it harder for the current model to fully adapt to recent changes.
As a result, enforcing consensus with too many snapshots over-regularizes the model and degrades reconstruction quality.

These results suggest that temporal consensus should prioritize recency rather than aggregating long histories, and that a small number of carefully weighted snapshots is sufficient for stabilizing continual neural mapping without impeding adaptation.

\end{document}